\documentclass[runningheads]{llncs}
\usepackage[T1]{fontenc}
\usepackage{graphicx,verbatim}
\usepackage{amsmath}
\usepackage{amssymb}
\usepackage{amsfonts}
\usepackage{multirow}
\usepackage{array}
\usepackage{booktabs}
\usepackage{colortbl}
\usepackage{xcolor}
\usepackage{amssymb}
\usepackage{makecell}
\usepackage{pifont}

\usepackage{placeins}

\usepackage{booktabs}
\usepackage{subcaption}
\usepackage{threeparttable}
\usepackage{siunitx}
\usepackage{xspace}
\newcommand{\ourmethod}{\texttt{CPAgents}\xspace}

\newif\ifrevision
\revisiontrue
\newcommand{\rev}[1]{%
  \ifrevision
    #1
  \else
    #1%
  \fi
}

\begin{document}

\title{CPAgents: Agentic Composite Phenotype Generation for Cardiac Disease Association}

\titlerunning{\rev{CPAgents for Disease-Wide Cardiac Phenotype Association}}

\newcommand{\equalcontrib}{\textsuperscript{*}}
\newcommand{\corrauthor}{\textsuperscript{$\dagger$}}

\author{
Zuoou Li\inst{1}\equalcontrib,
Wenlong Zhao\inst{2}\equalcontrib,
Kelly Yu\inst{1}\equalcontrib,
Weitong Zhang\inst{3},\\
Paul M. Matthews\inst{4,7,8},
Wenjia Bai\inst{3,4,5},
Bernhard Kainz\inst{3,6},
Mengyun Qiao\inst{1}\corrauthor
}

\authorrunning{Z. Li et al.}

\institute{
Department of Mechanical Engineering, University College London, London, UK
\and
CSIG Group, Tencent, Beijing, China
\and
Department of Computing, Imperial College London, London, UK
\and
Department of Brain Sciences, Imperial College London, London, UK
\and
Data Science Institute, Imperial College London, London, UK
\and
FAU Erlangen--Nürnberg, Erlangen, DE
\and
UK Dementia Research Institute, Imperial College London, London, UK
\and
Rosalind Franklin Institute, Harwell Science and Innovation Campus, Didcot, UK\\
\email{lizoleo9344@gmail.com, m.qiao@ucl.ac.uk}
}
  
\maketitle            
\begin{abstract}
Identifying robust associations between cardiac imaging phenotypes and clinical diseases is fundamental to population-scale cardiovascular research and reliable risk stratification. 
However, current phenome-wide association studies rely on pre-defined, single-variable phenotypes or expert-crafted features, which limits their ability to capture clinically meaningful non-linear effects and cross-phenotype interactions.
To address this, we propose \textbf{\ourmethod}, an iterative phenotype-\textbf{C}omposition framework for cardiovascular \textbf{P}henome-wide association study (PheWAS) that automatically constructs and validates interpretable composite phenotypes (e.g., polynomial, ratio, and interaction forms) from base imaging features.
Specifically, our system coordinates three agents: 
(i) an \textit{Analyst }that identifies statistical pathologies and nominates candidate transformations; 
(ii) a \textit{Proposer} that generates constrained, medically and statistically motivated expressions under numerical safety rules; and 
(iii) a \textit{Verifier} that evaluates candidates using multi-stage criteria and produces transparent evidence trails for accepted phenotypes.
Evaluated on a population-scale cardiac imaging cohort, the discovered composite phenotypes markedly improve disease discrimination: across 72 classifier–disease–metric combinations, our variants achieve the top rank in 56 cases versus 18 for baselines, with gains observed across all nine clinical disease categories.
Our framework yields compact, clinically interpretable phenotype formulas with transparent evidence trails, enabling scalable discovery of stronger phenotype-disease associations beyond expert-driven feature selection.


\end{abstract}
\section{Introduction}
\label{sec:intro}
Identifying robust associations between cardiac imaging phenotypes and clinical diseases is central to understanding disease mechanisms and building reliable risk stratification models \cite{skelly2012assessing}. Modern cardiovascular imaging cohorts, such as large-scale CMR and echocardiography biobanks, provide rich quantitative phenotypes that describe chamber size and function, ventricular mass and strain, aortic geometry \cite{ruijsink2020fully}. These imaging-derived traits serve as intermediate phenotypes linking genetic and environmental factors to clinical outcomes, and are increasingly used in mechanistic studies, biomarker discovery, and treatment response assessment   \cite{haupt2025explainable}.

The emergence of population-based cohorts with standardized CMR and echocardiography protocols, coupled with automated image analysis pipelines, has produced large imaging datasets at unprecedented scale \cite{bustin2020compressed}. While this enables systematic mapping of the cardiac imaging phenome to diverse clinical conditions and risk factors \cite{schuijf2005cardiac}, it also introduces substantial analytic challenges. In practice, the sheer number, redundancy, and complex dependency structure of imaging-derived traits create a substantial analytic bottleneck: it is non-trivial to decide which combinations of phenotypes to test, how to capture potentially non-linear relationships between imaging traits and clinical outputs, and how to ensure that resulting associations are robust and clinically interpretable \cite{luscher2024artificial}.

Previous work has largely adopted PheWAS-style pipelines \cite{denny2010phewas}, in which domain experts predefine a fixed set of single-variable imaging phenotypes and evaluate their associations with a range of diseases under carefully specified statistical models \cite{bai2020population,allen2024prospective} or machine learning algorithms \cite{huda2021machine,mukherjee2022confounding}. Subsequent work constructs composite indices based on prior clinical knowledge, for example, by combining related volumetric measures or summarizing regional features into global scores \cite{zhang2025multi,lan2025autoqual}. While these approaches are transparent and clinically grounded, they do not scale well without substantial manual effort, and are constrained by expert-defined features and linear modeling assumptions.

More recent studies have explored automated feature construction and data-driven discovery pipelines, including symbolic composition and agent-assisted data science systems \cite{abhyankar2025llm,wang2025medagent,fallahpour2025medrax}. Although such methods broaden the candidate feature space, they are typically developed for general predictive modelling rather than large-scale phenotype discovery with explicit confounder control and numerical safeguards. Greater flexibility can enhance expressiveness, but it may also introduce instability when applied to correlated imaging traits.

Both expert-driven and automated approaches face two major limitations in high-dimensional, population-scale imaging data. First, predefined feature sets restrict the ability to capture clinically meaningful non-linear effects and interactions across phenotypes. Second, expanding the feature space through adaptive composition inflates the multiple-testing burden and increases vulnerability to unstable findings under cohort heterogeneity and residual confounding.

To address these limitations, we propose \ourmethod, a \textbf{C}omposite-\textbf{P}henotype discovery framework for cardio-PheWAS that automatically constructs and validates interpretable composite imaging phenotypes capable of capturing non-linear effects and cross-phenotype interactions. The main contributions are:
(i) \textbf{Agentic iterative phenotype discovery.} We formulate phenotype discovery as an iterative analyse-propose-verify loop coordinated by specialised agents, enabling automatic construction of composite phenotypes within a constrained operator set.
(ii) \textbf{Statistics-informed composition and validation.} We first summarizes key statistical structure in the data (e.g., distributions, redundancy, and correlation patterns), and then performs principled screening and verification using cross-validated metrics. Candidate composites are retained only when improvements are consistent across folds and models and supported by statistical tests, mitigating selection-induced overfitting.
(iii) \textbf{Interpretable and reproducible outputs.} We represent each discovered composite phenotype as a compact, closed-form formula with a clear construction trace and cross-validated evidence summaries, enabling transparent auditing and facilitating reproducible downstream association analyses and external replication.

\begin{figure}
    \centering
    \includegraphics[width=0.99\linewidth]{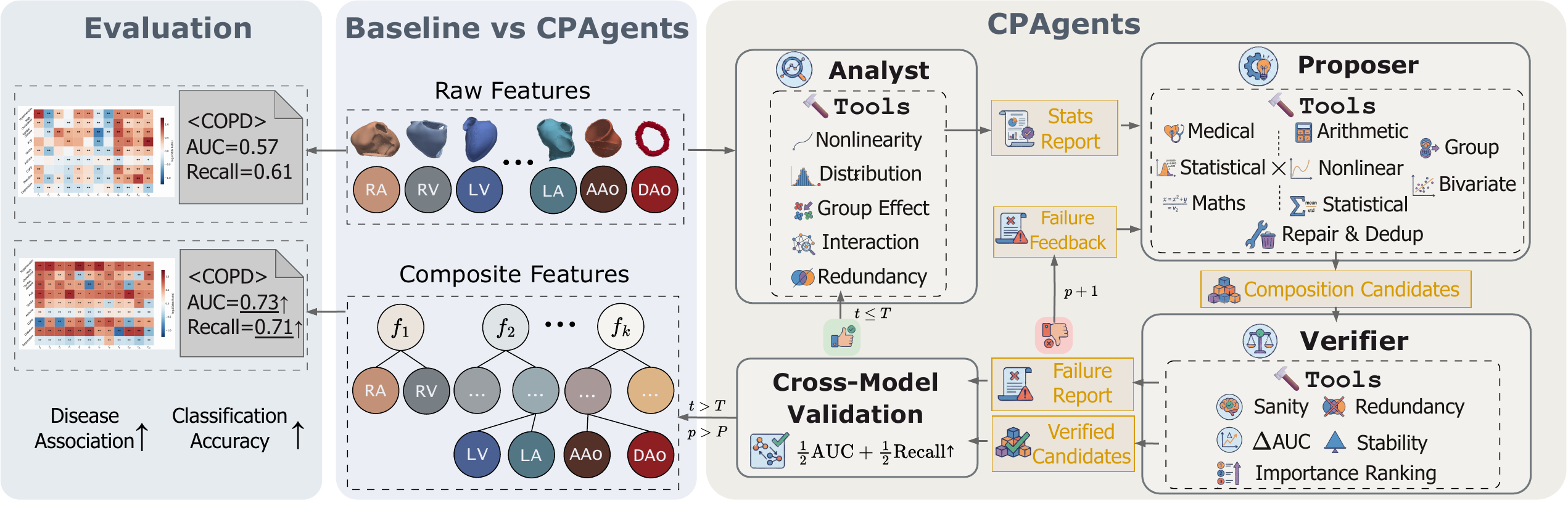}
    \caption{\textbf{Overview of \ourmethod, an agentic phenotype composition framework.} \ourmethod iteratively transforms raw cardiovascular phenotypes into highly predictive, hierarchical composite features ($f_1...f_k$). An \textit{Analyst} first profiles feature statistics to guide a \textit{Proposer}, which synthesizes candidate features using medical, statistical, and exploratory operations. Next, a \textit{Verifier} filters candidates via sanity, stability, and $\Delta$AUC checks. Candidates then undergo cross-model validation; failures trigger a feedback loop for refinement, while successes are retained. Ultimately, these transparent composite features significantly enhance downstream disease association mapping and classification accuracy.}
    \label{fig:overview}
\end{figure}

\section{Method}
\label{sec:method}

\textbf{Problem Setup.}
\label{sec:setup}
We consider a supervised \textit{composite phenotype} discovery problem for cardio--PheWAS across multiple target diseases.
Let $\mathcal{D}=\{(\mathbf{x}_i, y_i)\}_{i=1}^{N}$ denote a cohort of $N$ subjects. Each subject is represented by a comprehensive feature vector $\mathbf{x}_i \in \mathbb{R}^{d}$ (where $d = p+q$), comprising $p$ base cardiac imaging phenotypes (e.g., atrial/ventricular measures for \texttt{RA}/\texttt{RV}/\texttt{LA}/\texttt{LV} and aortic measures for \texttt{AAo}/\texttt{DAo}) and $q$ potential confounders (e.g., age, sex, alcohol intake, blood pressure). The target $y_i\in\{1,\ldots,K\}$ is a $K$-class disease label.

Our goal is to automatically discover and validate a compact set of interpretable composite phenotypes $\mathcal{F}=\{f_k\}_{k=1}^{M}$. Each $f_k$ is a closed-form mapping $f_k: \mathbb{R}^{d} \to \mathbb{R}$ operating on $\mathbf{x}$, such that $f_k \in \mathcal{G}$, where $\mathcal{G}$ denotes the space of valid expressions generated by a predefined operator set (e.g., clinical, statistical, and arithmetic transformations). These derived phenotypes aim to improve the predictive performance of downstream disease classification.

\textbf{Composite Phenotype Space.}
While standard linear or shallow models often under-parameterize non-linear effects, higher-order interactions, and group-dependent distributions, composite phenotypes make these underlying structures explicit in a compact, one-dimensional representation.

To formalize the search space $\mathcal{G}$, we define the generation of a composite phenotype as the construction of an Abstract Syntax Tree (AST), denoted as $\mathcal{T}$. Let $\mathcal{V} = \{x^{(1)}, x^{(2)}, \dots, x^{(d)}\}$ denote the set of all $d$ available scalar features from the input space. In this representation: 
(i) {Leaf Nodes ($\mathcal{L}$)} correspond either to a raw scalar feature $x^{(j)} \in \mathcal{V}$ or to domain-specific numerical constants $c \in \mathbb{R}$ organically generated by the Large Language Model (LLM).
(ii) {Internal Nodes ($\mathcal{N}$)} represent allowable operators $o \in \mathcal{O}$ applied to their child nodes.

To guarantee the validity, executability, and safety of the generated expressions on cross-sectional tabular data, the operator space $\mathcal{O}$ is strictly constrained to a predefined set of functions. These operators can be broadly categorized into:
(i) {Element-wise Non-linearities \& Transformations:} Mathematical functions (\textit{e.g.}, \texttt{log1p}, \texttt{sqrt}, \texttt{clip}, \texttt{power}, \texttt{where}) designed to handle skewed distributions and outlier masking.
(ii) {Arithmetic Interactions:} Basic algebraic operations (\textit{e.g.}, \texttt{add}, \texttt{sub}, \texttt{mul}, \texttt{div}) to construct interaction terms, ratios, and differences.
(iii) {Group-aware Reductions \& Statistics:} Advanced functions (\textit{e.g.}, \texttt{groupby}, \texttt{transform(mean/std)}, \texttt{rank}) that allow features to be normalized or evaluated relative to specific demographic or clinical sub-populations (\textit{e.g.}, stratifying by \texttt{Sex}).
Numerical safety guardrails, such as adding a smoothing term ($\epsilon = 10^{-9}$) to the denominator of all division operators (\texttt{div}), are strictly enforced at the AST parsing level to prevent zero-division.

\textbf{Clinical Example.} Coupled with the LLM's medical priors, this AST space enables the zero-shot discovery of complex, actionable indices. For example, the system can autonomously formulate demographic-adjusted phenotypes like:
\begin{equation}
    f_{\text{new}} = \frac{ \text{LVM}_{\text{index}} - \mu(\text{LVM}_{\text{index}} \mid \text{Sex}) }{ \sigma(\text{LVM}_{\text{index}} \mid \text{Sex}) + \epsilon }, \text{LVM}_{\text{index}} = \frac{\text{LVM}}{(\text{Height}/100)^{2.7} + \epsilon}
\end{equation}
This AST represents the \emph{sex-specific Z-score of allometrically indexed Left Ventricular Mass}, where $\mu$ and $\sigma$ denote group-conditional statistics. Crucially, domain-specific constants like $2.7$ (a cardiologic allometric scaling factor) and $100$ are elicited directly from the LLM's internal knowledge, bypassing computationally expensive random search. Such explicit, domain-aware features significantly enhance the performance of interpretable downstream classifiers.

\textbf{Iterative Search Procedure.}
We formulate automated feature engineering as an iterative search over a constrained composite-phenotype space, orchestrated by a closed-loop system of three specialized agents: an \emph{Analyst}, a \emph{Proposer}, and a \emph{Verifier}.
At each iteration $(t)$, the system seeks to augment the current feature set $\mathcal{F}^{(t)}$ through a structured four-phase procedure: \emph{statistical profiling}, \emph{transformation generation}, \emph{multi-stage filtering}, and \emph{global acceptance testing}.

\textbf{Phase 1: Statistical Profiling.}
To prevent blind combinatorial search, the Analyst Agent systematically profiles $\mathcal{F}^{(t)}$ to uncover data characteristics. This yields five comprehensive statistical reports: 
(i) {Univariate shape assessment} via $\Delta \text{AIC}$ between linear and spline logistic models to detect non-linear and monotonicity benefits; 
(ii) {Distributional characterization} through skewness, kurtosis, and outlier rates; 
(iii) {Pairwise interaction screening} using mutual information, followed by cross-validated logistic confirmation; 
(iv) {Group-effect analysis} via intraclass correlation (ICC) and Cramér's V to assess safe group-level standardizations; and 
(v) {Redundancy clustering} on the absolute correlation matrix. 
Subsequently, a Large Language Model (LLM) compresses these statistical diagnostics into structured, per-feature recommendations, prioritizing interaction pairs and safely bounded group operations.

\textbf{Phase 2: Transformation Generation.}
Guided by the structured hints from the \emph{Analyst}, the \emph{Proposer} generates $C = \min(|\mathcal{F}^{(t)}|, C_{\max})$ new composite feature candidates $f_{\text{cand}} \in \mathcal{G}$. To maintain a balance between domain validity and empirical discovery, these candidates are produced via three separate LLM calls, corresponding to medical, statistical, and exploratory priorities allocated in a 2:2:1 ratio:
(i) {Medical Priors (40\%):} Features derived purely from the LLM's internal clinical knowledge (\textit{e.g.}, specific risk indices or established physiological ratios), irrespective of statistical hints.
(ii) {Statistical Transformations (40\%):} Features directly addressing the \emph{Analyst}'s reports, such as applying \texttt{log1p} to heavy-tailed distributions or creating interaction terms with a high $\Delta \text{AIC}$.
(iii) {Exploratory Mathematics (20\%):} Safe, automated mathematical combinations within the operator space $\mathcal{O}$, designed to discover latent, unintuitive patterns.

Each LLM call is conditioned on the current feature set $\mathcal{F}^{(t)}$, forbidden target labels $y$, safe group operations, and the \emph{Analyst}'s per-feature recommendations, interaction pairs, and redundancy clusters. All generated expressions are constrained to an interpreter that performs AST-level validation, restricting operations to the predefined operator space $\mathcal{O}$. Failed candidates are re-prompted to the LLM with error messages for up to two repair attempts.

\textbf{Phase 3: Multi-Stage Filtering.}
To distill the top-ranked proposals, the \textit{Verifier} evaluates the newly proposed and existing features through a multi-stage screening: 
(i) {Marginal Utility Check:} Evaluating the isolated $\Delta \text{AUC}$ of each new feature.
(ii) {Stability Selection:} Applying ElasticNet \cite{zou2005regularization} regularization to penalize spurious correlations and verify stability.
(iii) {Deduplication:} Removing redundant features via a combination of statistical correlation clustering and AST-based expression textual matching.
(iv) {Global Importance Assessment:} Training a LightGBM model and utilizing SHapley Additive exPlanations (SHAP)~\cite{lundberg2017unified} values to extract the absolute top-ranked impactful features.

\textbf{Phase 4: Global Acceptance Criteria.}
Finally, the top-ranked candidate features, denoted as $\mathcal{F}_{\text{cand}}$, undergo empirical validation. To assess their downstream predictive utility, we define a composite evaluation metric $\mathcal{S} = 0.5 \cdot \text{AUC} + 0.5 \cdot \text{Recall}$. The augmented feature space, $\mathcal{F}^{(t)} \cup \mathcal{F}_{\text{cand}}$, is evaluated on a hold-out validation dataset using a panel of four base classifiers (SVM, LDA, AdaBoost, and MLP). The feature update is governed by a ensemble agreement threshold: if $\mathcal{S}(\mathcal{F}^{(t)}\cup \mathcal{F}_{\text{cand}})$ improves over the previous $\mathcal{S}(\mathcal{F}^{(t)})$ on at least two out of the four models, the new features are accepted if $t<$ max iteration $T$, yielding $\mathcal{F}^{(t+1)} = \mathcal{F}^{(t)} \cup \mathcal{F}_{\text{cand}}$. Otherwise, the candidates are discarded ($\mathcal{F}^{(t+1)} = \mathcal{F}^{(t)}$), and the \textit{Verifier} passes the performance feedback back to the \textit{Proposer} to refine its output for the next iteration. To bound the computational cost, we uses early stopping, terminating after $P=3$ consecutive rejections.

\section{Experiments and Results}

\textbf{Dataset and Settings.} We evaluate the proposed framework on the UK Biobank dataset\footnote{http://www.ukbiobank.ac.uk/register-apply}, a population-scale cardiac imaging cohort comprising 26,893 participants. We consider $K=9$ clinical disease categories, including hypertension, high cholesterol, cardiac disease, PVD, stroke, asthma, COPD, diabetes, and depression. Cardiac imaging phenotypes are extracted using a standardized CMR analysis pipeline~\cite{bai2020population} and released by the UK Biobank\footnote{https://biobank.ndph.ox.ac.uk/showcase/label.cgi?id=157}. The following confounders are included in all analyses: sex, age, weight, height, alcohol intake, alcohol intake log10, systolic blood pressure, and diastolic blood pressure.

\textbf{Evaluation Metrics.} We divide the dataset into training, validation, and test sets with a ratio of 6:2:2. Composite phenotype discovery, model selection, and threshold tuning are performed exclusively on the training and validation sets to prevent information leakage. Final evaluation is conducted once on the independent test set. We report the Area Under the ROC Curve (AUC) and Recall as primary performance metrics. Additionally, to interpret why the composite features generated by our method are beneficial for classification, we utilize the Silhouette score. For a given sample $i$, it is defined as $s(i) = (b(i) - a(i)) / \max\{a(i), b(i)\}$, where $a(i)$ represents the mean intra-class distance to all other samples within the same class, and $b(i)$ is the mean inter-class distance to the samples of the nearest neighboring class. The overall score is the average of $s(i)$ across all samples, ranging from $-1$ to $1$.

\textbf{Implementation Details.}
The \textit{Analyst} and \textit{Proposer} agents use DeepSeek-V3.2~\cite{liu2025deepseek} as the LLM backend via LangChain with the default temperature for a balance between deterministic reasoning and exploration. During the $\Delta\text{AIC}$ profiling phase, the \textit{Analyst} models non-linear effects using B-splines (degrees of freedom $df=4$, polynomial degree $d=3$) with five-fold cross-validation. Clinical definitions of base features (\textit{e.g.}, \texttt{LVEDV}: Left Ventricular End-Diastolic Volume) are injected into the \textit{Proposer} system prompts to ground candidate generation. The \textit{Verifier} performs stability selection with 50 bootstrap runs of an ElasticNet classifier ($\ell_1$-ratio $=0.5$, inverse regularization $C=0.5$). To prevent data leakage, all feature engineering and selection are restricted to the training set, and final performance is reported on held-out validation and test sets. Each search uses a fixed random seed for reproducibility, and all metrics are averaged over five independent runs with different seeds to ensure robustness.

\textbf{Baselines.} To demonstrate the benefits of our composite features, we employ several baselines: expert-defined features from~\cite{bai2020population}, features selected by MESHAgent~\cite{zhang2025multi} through debate and consensus. We employ ChatGPT and DeepSeek, two cutting-edge LLMs accessed via the official APIs. All experiments are performed on a single NVIDIA RTX 6000 GPU.

\begin{figure*}[ht]
  \centering
  \begin{subfigure}[t]{0.49\textwidth}
    \centering
    \includegraphics[width=\linewidth]{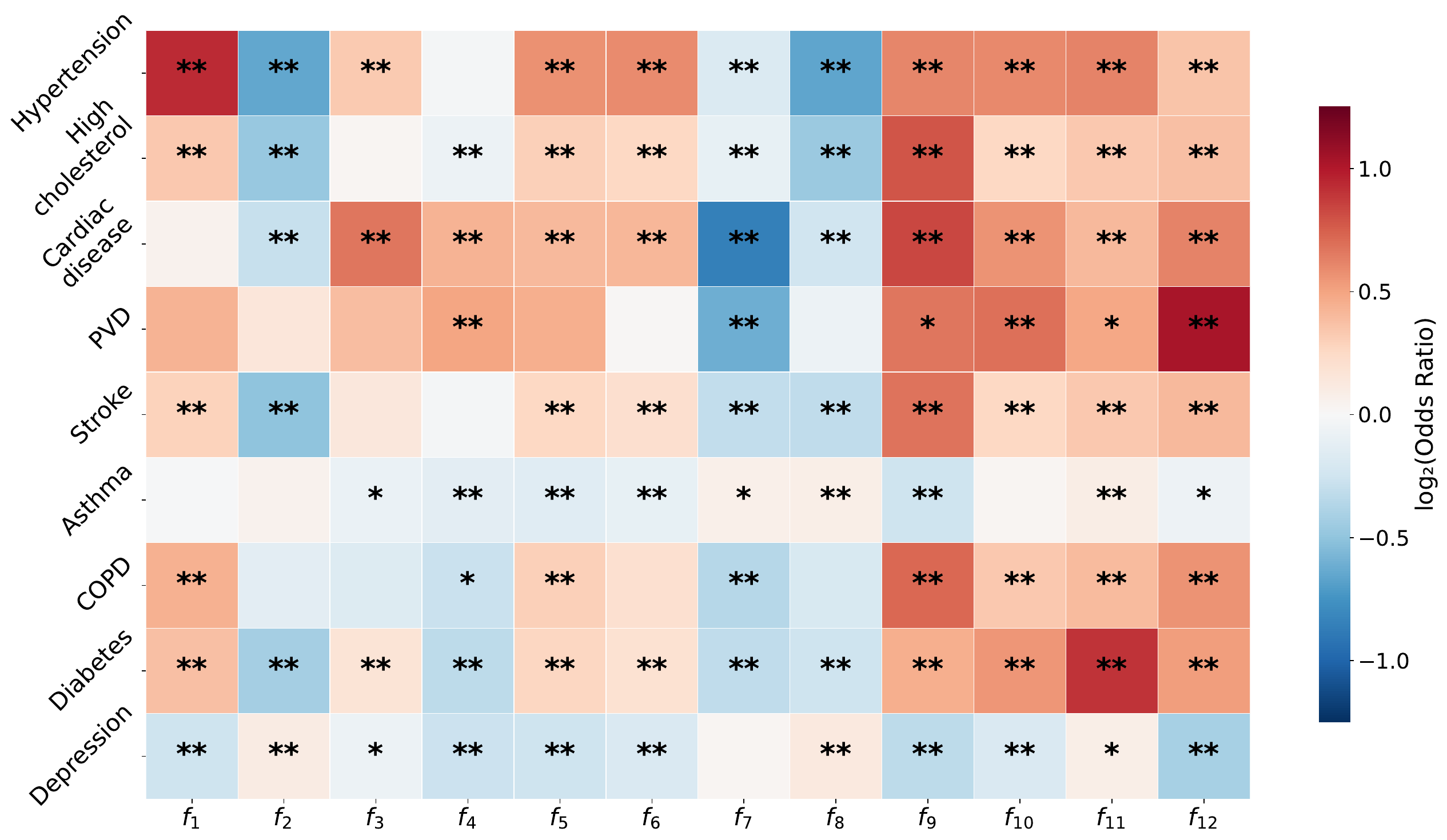}
    \label{fig:heatmap_expert}
  \end{subfigure}\hfill
  \begin{subfigure}[t]{0.49\textwidth}
    \centering
    \includegraphics[width=\linewidth]{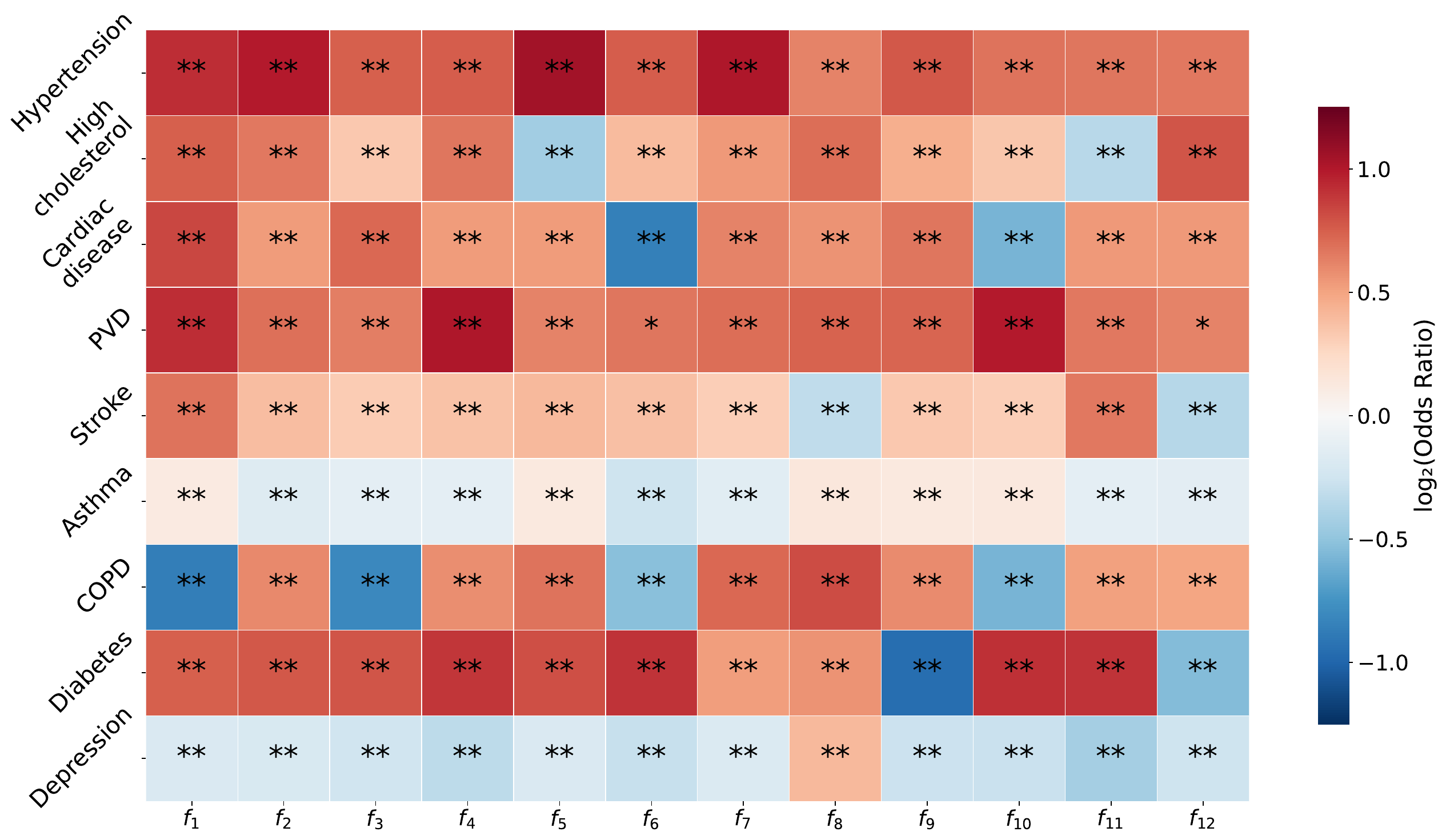}
    \label{fig:heatmap_ours}
  \end{subfigure}
  \caption{Disease--phenotype association heatmaps for expert-defined features (left) and our composite phenotypes (right). Except for asthma and depression, composite phenotypes produce more concentrated and consistent association patterns across diseases, supporting improved cluster-level interpretability of the learned features.}
  \label{fig:heatmap_compare}
\end{figure*}

\textbf{Results.} 
(i) Disease Association Study. The disease--phenotype association heatmap (Figure~\ref{fig:heatmap_compare}) shows that composite phenotypes discovered by our framework yield clearer, more concentrated association patterns compared to base and expert-defined traits, with stronger effects in key cardiac and other disease groups.
(ii) \textit{Diagnostic performance.} Table~\ref{tab:main_result} compares disease discrimination using our composite phenotypes with expert-selected single features~\cite{bai2020population} and MESHAgents-based features~\cite{zhang2025multi} across four classifiers and nine disease endpoints. A ranking analysis over all 72 classifier–disease–metric combinations shows that our two composite-phenotype variants occupy the top rank in \textbf{56} cases versus \textbf{18} for the baselines, and appear in the top two in \textbf{51} settings (compared with 33 and 31 for expert-defined and MESHAgents features), indicating consistently competitive and superior performance across diseases rather than gains concentrated in a few favourable tasks.

\newcommand{\method}[2]{\makecell[l]{#1\\[-4.2pt]{\small\texttt{#2}}}} 

\newcommand{\score}[2]{%
  \makecell[c]{%
    #1\\[-4.2pt]
    \makebox[0pt][c]{\scriptsize \quad $\pm#2$}%
  }%
}

\begin{table*}[ht]
\centering
\caption{Performance comparison (AUC and Recall) across classification models and diseases. The \textbf{best} results in each block are highlighted in boldface.}
\label{tab:main_result}
\resizebox{\textwidth}{!}{
\begin{tabular}{l *{18}{c}}
\toprule
\multirow{2}{*}{\textbf{Methods}}
& \multicolumn{2}{c}{\textbf{Hypertension}}
& \multicolumn{2}{c}{\textbf{High cholesterol}}
& \multicolumn{2}{c}{\textbf{Cardiac disease}}
& \multicolumn{2}{c}{\textbf{PVD}}
& \multicolumn{2}{c}{\textbf{Stroke}}
& \multicolumn{2}{c}{\textbf{Asthma}}
& \multicolumn{2}{c}{\textbf{COPD}}
& \multicolumn{2}{c}{\textbf{Diabetes}}
& \multicolumn{2}{c}{\textbf{Depression}} \\
\cmidrule(lr){2-3}\cmidrule(lr){4-5}\cmidrule(lr){6-7}\cmidrule(lr){8-9}\cmidrule(lr){10-11}\cmidrule(lr){12-13}\cmidrule(lr){14-15}\cmidrule(lr){16-17}\cmidrule(lr){18-19}
& \textbf{AUC}  $\uparrow$ & \textbf{Recall} $\uparrow$
& \textbf{AUC} $\uparrow$ & \textbf{Recall} $\uparrow$
& \textbf{AUC} $\uparrow$  & \textbf{Recall} $\uparrow$
& \textbf{AUC} $\uparrow$  & \textbf{Recall} $\uparrow$
& \textbf{AUC} $\uparrow$  & \textbf{Recall} $\uparrow$
& \textbf{AUC} $\uparrow$  & \textbf{Recall}$\uparrow$
& \textbf{AUC} $\uparrow$ & \textbf{Recall} $\uparrow$
& \textbf{AUC} $\uparrow$  & \textbf{Recall} $\uparrow$
& \textbf{AUC} $\uparrow$  & \textbf{Recall} $\uparrow$  \\
\toprule

\multicolumn{19}{c}{\cellcolor{gray!20}\textit{SVM}}\\
\midrule
Experts~\cite{bai2020population} 
& \score{0.755}{0.007} & \score{\underline{0.779}}{0.009} & \score{0.663}{0.014} & \score{0.683}{0.026} & \score{0.737}{0.010} & \score{0.620}{0.033} & \score{0.710}{0.075} & \score{\textbf{0.783}}{0.183} & \score{0.635}{0.040} & \score{\underline{0.615}}{0.070} & \score{0.566}{0.011} & \score{0.539}{0.031} & \score{\underline{0.720}}{0.052} & \score{\underline{0.686}}{0.096} & \score{0.781}{0.008} & \score{0.715}{0.028} & \score{0.601}{0.007} & \score{\underline{0.566}}{0.045} \\
\texttt{MESHAgents}~\cite{zhang2025multi} 
& \score{0.748}{0.009} & \score{0.771}{0.013} & \score{\underline{0.664}}{0.014} & \score{\textbf{0.689}}{0.025} & \score{0.734}{0.009} & \score{0.622}{0.027} & \score{0.703}{0.049} & \score{\underline{0.710}}{0.133} & \score{0.623}{0.047} & \score{0.549}{0.060} & \score{0.565}{0.009} & \score{0.550}{0.023} & \score{0.712}{0.063} & \score{\textbf{0.692}}{0.110} & \score{0.771}{0.008} & \score{\underline{0.717}}{0.022} & \score{0.603}{0.009} & \score{\textbf{0.586}}{0.016} \\
\midrule
\method{Ours}{GPT-5.1} 
& \score{\underline{0.757}}{0.006} & \score{0.775}{0.009} & \score{\textbf{0.667}}{0.014} & \score{0.679}{0.018} & \score{\textbf{0.752}}{0.009} & \score{\textbf{0.646}}{0.030} & \score{\textbf{0.713}}{0.079} & \score{0.603}{0.199} & \score{\underline{0.641}}{0.035} & \score{0.609}{0.033} & \score{\textbf{0.582}}{0.012} & \score{\textbf{0.567}}{0.021} & \score{\textbf{0.726}}{0.040} & \score{0.654}{0.100} & \score{\textbf{0.785}}{0.004} & \score{\textbf{0.736}}{0.015} & \score{\underline{0.607}}{0.013} & \score{0.564}{0.016} \\
\method{Ours}{DeepSeek V3.2} 
& \score{\textbf{0.759}}{0.006} & \score{\textbf{0.780}}{0.006} & \score{0.663}{0.012} & \score{\underline{0.687}}{0.039} & \score{\underline{0.746}}{0.011} & \score{\underline{0.634}}{0.037} & \score{\underline{0.712}}{0.088} & \score{0.553}{0.183} & \score{\textbf{0.652}}{0.040} & \score{\textbf{0.632}}{0.078} & \score{\underline{0.580}}{0.009} & \score{\underline{0.565}}{0.015} & \score{0.706}{0.058} & \score{0.641}{0.100} & \score{\underline{0.784}}{0.005} & \score{0.725}{0.016} & \score{\textbf{0.609}}{0.014} & \score{0.560}{0.023} \\

\midrule
\multicolumn{19}{c}{\cellcolor{gray!20}\textit{AdaBoost}}\\
\midrule
Experts~\cite{bai2020population} 
& \score{0.741}{0.010} & \score{0.699}{0.018} & \score{\underline{0.667}}{0.006} & \score{\underline{0.655}}{0.027} & \score{\underline{0.728}}{0.008} & \score{0.637}{0.023} & \score{0.626}{0.175} & \score{\underline{0.667}}{0.236} & \score{0.615}{0.028} & \score{0.561}{0.037} & \score{0.563}{0.012} & \score{\textbf{0.589}}{0.043} & \score{0.651}{0.049} & \score{0.604}{0.082} & \score{\underline{0.772}}{0.006} & \score{0.687}{0.018} & \score{0.619}{0.006} & \score{0.588}{0.017} \\
\texttt{MESHAgents}~\cite{zhang2025multi} 
& \score{0.738}{0.011} & \score{0.695}{0.016} & \score{\textbf{0.669}}{0.007} & \score{\textbf{0.658}}{0.022} & \score{0.724}{0.009} & \score{\textbf{0.657}}{0.014} & \score{\underline{0.682}}{0.074} & \score{\textbf{0.817}}{0.207} & \score{0.611}{0.026} & \score{\underline{0.565}}{0.060} & \score{\underline{0.569}}{0.015} & \score{\underline{0.582}}{0.044} & \score{0.658}{0.041} & \score{0.567}{0.092} & \score{0.770}{0.010} & \score{\underline{0.694}}{0.019} & \score{\underline{0.620}}{0.008} & \score{\underline{0.594}}{0.017} \\
\midrule
\method{Ours}{GPT-5.1} 
& \score{\underline{0.752}}{0.008} & \score{\underline{0.725}}{0.016} & \score{\underline{0.667}}{0.007} & \score{0.638}{0.023} & \score{\textbf{0.732}}{0.009} & \score{\underline{0.647}}{0.020} & \score{0.602}{0.154} & \score{\underline{0.577}}{0.211} & \score{\underline{0.628}}{0.015} & \score{\textbf{0.588}}{0.043} & \score{\textbf{0.583}}{0.004} & \score{\underline{0.565}}{0.025} & \score{\textbf{0.733}}{0.036} & \score{\textbf{0.713}}{0.067} & \score{\underline{0.785}}{0.013} & \score{\underline{0.713}}{0.027} & \score{\underline{0.621}}{0.004} & \score{\underline{0.601}}{0.009} \\
\method{Ours}{DeepSeek V3.2} 
& \score{\textbf{0.754}}{0.006} & \score{\textbf{0.733}}{0.015} & \score{\textbf{0.669}}{0.010} & \score{\underline{0.648}}{0.028} & \score{0.726}{0.013} & \score{0.641}{0.027} & \score{\textbf{0.684}}{0.187} & \score{0.560}{0.339} & \score{\textbf{0.639}}{0.029} & \score{\underline{0.569}}{0.057} & \score{\underline{0.574}}{0.013} & \score{0.550}{0.021} & \score{\underline{0.700}}{0.049} & \score{\underline{0.654}}{0.095} & \score{\textbf{0.789}}{0.007} & \score{\textbf{0.724}}{0.006} & \score{\textbf{0.624}}{0.011} & \score{\textbf{0.606}}{0.041} \\

\midrule
\multicolumn{19}{c}{\cellcolor{gray!20}\textit{MLP}}\\
\midrule
Experts~\cite{bai2020population} 
& \score{\underline{0.761}}{0.006} & \score{\textbf{0.775}}{0.017} & \score{\textbf{0.663}}{0.013} & \score{\underline{0.701}}{0.031} & \score{\underline{0.738}}{0.008} & \score{\underline{0.631}}{0.028} & \score{0.628}{0.133} & \score{0.643}{0.217} & \score{\underline{0.595}}{0.018} & \score{\underline{0.621}}{0.166} & \score{\underline{0.564}}{0.008} & \score{0.509}{0.031} & \score{0.669}{0.010} & \score{0.612}{0.149} & \score{\underline{0.784}}{0.012} & \score{\underline{0.705}}{0.030} & \score{0.594}{0.010} & \score{0.591}{0.006} \\
\texttt{MESHAgents}~\cite{zhang2025multi} 
& \score{0.754}{0.010} & \score{\underline{0.768}}{0.015} & \score{0.658}{0.012} & \score{0.686}{0.019} & \score{0.728}{0.008} & \score{0.597}{0.020} & \score{\underline{0.640}}{0.042} & \score{\underline{0.660}}{0.159} & \score{0.576}{0.036} & \score{0.577}{0.144} & \score{0.555}{0.006} & \score{0.470}{0.060} & \score{\underline{0.673}}{0.024} & \score{\underline{0.662}}{0.091} & \score{0.772}{0.012} & \score{0.696}{0.039} & \score{\underline{0.598}}{0.005} & \score{\underline{0.599}}{0.039} \\
\midrule
\method{Ours}{GPT-5.1} 
& \score{\textbf{0.763}}{0.007} & \score{0.770}{0.020} & \score{\underline{0.662}}{0.015} & \score{\underline{0.712}}{0.017} & \score{0.735}{0.010} & \score{\textbf{0.658}}{0.021} & \score{\underline{0.637}}{0.150} & \score{0.643}{0.286} & \score{\textbf{0.619}}{0.022} & \score{0.563}{0.108} & \score{0.545}{0.008} & \score{\underline{0.521}}{0.044} & \score{0.616}{0.095} & \score{0.546}{0.138} & \score{\textbf{0.791}}{0.013} & \score{\textbf{0.716}}{0.035} & \score{\textbf{0.610}}{0.016} & \score{\underline{0.599}}{0.027} \\
\method{Ours}{DeepSeek V3.2} 
& \score{0.710}{0.007} & \score{0.716}{0.017} & \score{0.657}{0.010} & \score{\textbf{0.757}}{0.037} & \score{\textbf{0.740}}{0.013} & \score{0.629}{0.058} & \score{\textbf{0.741}}{0.091} & \score{\textbf{0.707}}{0.301} & \score{0.591}{0.062} & \score{\textbf{0.637}}{0.409} & \score{\textbf{0.565}}{0.009} & \score{\textbf{0.542}}{0.078} & \score{\textbf{0.730}}{0.064} & \score{\textbf{0.708}}{0.155} & \score{\underline{0.784}}{0.010} & \score{\underline{0.705}}{0.017} & \score{\underline{0.599}}{0.015} & \score{\textbf{0.663}}{0.017} \\

\midrule
\multicolumn{19}{c}{\cellcolor{gray!20}\textit{LDA}}\\
\midrule
Experts~\cite{bai2020population} 
& \score{\underline{0.760}}{0.006} & \score{0.699}{0.007} & \score{\textbf{0.675}}{0.010} & \score{0.637}{0.016} & \score{0.747}{0.008} & \score{\textbf{0.667}}{0.018} & \score{0.636}{0.142} & \score{0.610}{0.292} & \score{\underline{0.619}}{0.038} & \score{0.593}{0.063} & \score{0.583}{0.013} & \score{\textbf{0.571}}{0.020} & \score{\underline{0.712}}{0.041} & \score{\underline{0.644}}{0.044} & \score{\underline{0.792}}{0.007} & \score{\underline{0.716}}{0.019} & \score{\underline{0.619}}{0.010} & \score{0.589}{0.012} \\
\texttt{MESHAgents}~\cite{zhang2025multi} 
& \score{0.755}{0.009} & \score{0.701}{0.008} & \score{\underline{0.673}}{0.011} & \score{0.637}{0.017} & \score{0.740}{0.009} & \score{0.661}{0.023} & \score{\textbf{0.666}}{0.060} & \score{\textbf{0.630}}{0.178} & \score{0.612}{0.039} & \score{0.589}{0.072} & \score{0.583}{0.008} & \score{\underline{0.570}}{0.019} & \score{\textbf{0.730}}{0.035} & \score{\textbf{0.669}}{0.079} & \score{0.787}{0.007} & \score{0.705}{0.017} & \score{0.616}{0.009} & \score{\underline{0.590}}{0.018} \\
\midrule
\method{Ours}{GPT-5.1} 
& \score{\textbf{0.762}}{0.006} & \score{\textbf{0.728}}{0.006} & \score{0.671}{0.010} & \score{\underline{0.644}}{0.017} & \score{\underline{0.750}}{0.014} & \score{\underline{0.654}}{0.024} & \score{\underline{0.658}}{0.087} & \score{\underline{0.543}}{0.266} & \score{0.604}{0.031} & \score{\underline{0.595}}{0.060} & \score{\textbf{0.590}}{0.008} & \score{0.554}{0.019} & \score{0.705}{0.014} & \score{0.633}{0.079} & \score{\underline{0.792}}{0.006} & \score{\textbf{0.726}}{0.032} & \score{0.614}{0.012} & \score{\underline{0.590}}{0.028} \\
\method{Ours}{DeepSeek V3.2} 
& \score{\textbf{0.764}}{0.008} & \score{\underline{0.725}}{0.007} & \score{0.667}{0.008} & \score{\textbf{0.673}}{0.042} & \score{\textbf{0.751}}{0.015} & \score{\underline{0.656}}{0.034} & \score{0.529}{0.115} & \score{0.537}{0.210} & \score{\textbf{0.621}}{0.044} & \score{\textbf{0.598}}{0.064} & \score{\underline{0.588}}{0.013} & \score{\underline{0.559}}{0.024} & \score{0.678}{0.043} & \score{0.576}{0.068} & \score{\textbf{0.794}}{0.004} & \score{0.720}{0.023} & \score{\textbf{0.621}}{0.011} & \score{\textbf{0.593}}{0.027} \\

\bottomrule
\end{tabular}}
\end{table*}

\textbf{Robustness and Ablation Studies.}
As shown in Table~\ref{tab:ablation}, the full framework achieves the highest overall performance, yielding an average AUC of $0.686 \pm 0.076$ and a Recall of $0.643 \pm 0.069$. The ablation experiments identify the \textit{Verifier} as the most critical module in the framework; its removal causes the most severe performance degradation, plummeting the AUC to $0.590 \pm 0.006$ and Recall to $0.544 \pm 0.012$, followed by the \textit{Analyst}, Feedback, and SHAP Ranking.

\begin{table*}[ht]
\centering
\caption{Ablation and interpretability analysis. \textbf{Left:} Ablation studies of the proposed framework. \textbf{Right:} Cluster-level interpretability, showing that composite phenotypes improve classification (silhouette scores $\times 10$ for readability).}
\label{tab:ablation}

\small 

\begin{subtable}[t]{0.64\textwidth}
    \centering
    \begin{tabular}{lcc}
    \toprule
    \textbf{Setting} & \textbf{$\overline{\mathrm{AUC}}\pm\mathrm{std}$} & \textbf{$\overline{\mathrm{Recall}}\pm\mathrm{std}$} \\
    \midrule
    w/o Analyst      & 0.672$\pm$0.007 & 0.626$\pm$0.021 \\
    w/o Verifier     & 0.590$\pm$0.006 & 0.544$\pm$0.012 \\
    w/o Feedback     & 0.679$\pm$0.085 & 0.640$\pm$0.157 \\
    w/o SHAP Ranking & 0.685$\pm$0.011 & 0.637$\pm$0.016 \\
    \textbf{\ourmethod} & \textbf{0.686$\pm$0.076} & \textbf{0.643$\pm$0.069} \\
    \bottomrule
    \end{tabular}
\end{subtable}
\hfill
\begin{subtable}[t]{0.35\textwidth}
    \centering
    \begin{tabular}{lc}
    \toprule
    \textbf{Feature Set} & \textbf{$\overline{\mathrm{Silhouette}\times10}$} \\
    \midrule
    Raw Feature      & 0.293 \\
    Expert           & 0.276 \\
    MESHAgent        & 0.113 \\
    Constituent      & 0.094 \\
    \textbf{Composite} & \textbf{0.413} \\
    \bottomrule
    \end{tabular}
\end{subtable}

\end{table*}

\textbf{Cluster Separability.}
We employ the silhouette score to quantify cluster separability. As shown in Table~\ref{tab:ablation}, our composite phenotype attains the highest value, about \textbf{1.5}$\times$ that of the expert-defined feature set. The raw base feature has a silhouette score of \textbf{0.029}, which increases to \textbf{0.041} after applying our composition (an absolute gain of 0.012, roughly \textbf{41\%} relative), indicating improved class separation and helping to explain the downstream classification benefits of the learned composite phenotypes.

\section{Conclusion}
We have presented an agentic, iterative phenotype‐composition framework that discovers clinically interpretable composite cardiac imaging phenotypes and strengthens phenotype–disease association signals beyond expert-defined single traits. By combining analyst, proposer, and verifier agents with statistically grounded robustness checks, our approach yields compact, auditable phenotype libraries that are better aligned with underlying pathophysiology. Future work will extend the framework to multi-modal data to evaluate how composite phenotypes can further support clinical decision-making and trial design.

\begin{credits}
\subsubsection{\ackname}
Z.L. and M.Q. is supported by the Departmental Funding of the Department of Mechanical Engineering, University College London.
W.Z. is supported by the JADS programme and the UKRI Centre for Doctoral Training in AI for Healthcare (EP/S023283/1). 
P.M.M. acknowledges generous personal support from the Edmond J. Safra Foundation and Lily Safra, an NIHR Senior Investigator Award, Rosalind Franklin Institute, UKRI EPSRC funding, and UK Dementia Research Institute, which is funded predominantly by the UKRI Medical Research Council. 
W.B. is supported by EPSRC CVD-Net Programme Grant (EP/Z531297/1) and BHF New Horizons Grant (NH/F/23/70013). 
B.K. HPC resources from NHR@FAU (projects b143dc, b180dc), funded by federal and Bavarian state authorities and Gerhard Wellein's and his team's HPC approach. NHR@FAU hardware is partially funded by DFG 440719683. Additional support was received from ERC projects MIA-NORMAL 101083647, CHARMS 101246053, and ORACLE 101326871, DFG 513220538 and 512819079, and by the state of Bavaria (HTA). 
This research was conducted using the UK Biobank Resource under Application Number 18545. We thank all UK Biobank participants and staff.

\subsubsection{\discintname}
The authors have no competing interests to declare that are relevant to the content of this article.
\end{credits}
%
%
\bibliographystyle{vancouver}
\bibliography{ref}

\end{document}